\documentclass{article} 
\usepackage{iclr2026_conference,times}

\usepackage{amsmath,amsfonts,bm}

\def\eqref#1{equation~\ref{#1}}

\def\1{\bm{1}}

\DeclareMathAlphabet{\mathsfit}{\encodingdefault}{\sfdefault}{m}{sl}
\SetMathAlphabet{\mathsfit}{bold}{\encodingdefault}{\sfdefault}{bx}{n}

\usepackage{hyperref}
\usepackage{url}
\usepackage{booktabs}       
\usepackage{amsfonts}       
\usepackage{nicefrac}       
\usepackage{microtype}      
\usepackage{xcolor}         
  
\usepackage{amsmath}
\usepackage{graphicx}
\usepackage{listings}
\usepackage{amsfonts}
\usepackage{subcaption}
\usepackage{multirow}
\usepackage{colortbl}
\usepackage{tabularx}
\usepackage{booktabs}
\usepackage{float}
\usepackage{wrapfig}

\definecolor{mypurple}{RGB}{253,245,250}
\newcommand{\MODEL}{SophiaVL-R1}

\title{\MODEL: Reinforcing MLLMs Reasoning with Thinking Reward}

\author{%
 Kaixuan Fan\textsuperscript{1,2}\thanks{Equal contribution.}\;\;\thanks{This work was done during her internship at Shanghai Artifcial Intelligence Laboratory}\;,\; 
 Kaituo Feng\textsuperscript{1}\footnotemark[1]\;,\; Haoming Lyu\textsuperscript{2},\; 
 Dongzhan Zhou\textsuperscript{2}\footnotemark[3]\;,\; Xiangyu Yue\textsuperscript{1}\thanks{Corresponding Authors.} \\
 \\
 \textsuperscript{1}MMLab, The Chinese University of Hong Kong
\quad
\textsuperscript{2}Shanghai Artifcial Intelligence Laboratory \\
\\
\centerline{\url{https://github.com/kxfan2002/SophiaVL-R1}}
}

\iclrfinalcopy 
\begin{document}

\maketitle

\begin{abstract}

Recent advances have shown success in eliciting strong reasoning abilities in multimodal large language models (MLLMs) through rule-based reinforcement learning (RL) with outcome rewards. 
However, this paradigm typically lacks supervision over the thinking process leading to the final outcome. 
As a result, the model may learn sub-optimal reasoning strategies, which can hinder its generalization ability. 
In light of this, we propose SophiaVL-R1, as an attempt to add reward signals for the thinking process in this paradigm.
To achieve this, we first train a thinking reward model that evaluates the quality of the entire thinking process.
Given that the thinking reward may be unreliable for certain samples due to reward hacking, we propose the Trust-GRPO method, which assigns a trustworthiness weight to the thinking reward during training. 
This weight is computed based on the thinking reward comparison of responses leading to correct answers versus incorrect answers, helping to mitigate the impact of potentially unreliable thinking rewards.
Moreover, we design an annealing training strategy that gradually reduces the thinking reward over time, allowing the model to rely more on the accurate rule-based outcome reward in later training stages.
Experiments show that our SophiaVL-R1 surpasses a series of reasoning MLLMs on various benchmarks (\textit{e.g.}, MathVisita, MMMU), demonstrating strong reasoning and generalization capabilities. Notably, our SophiaVL-R1-7B even outperforms LLaVA-OneVision-72B on most benchmarks, despite the latter having $10\times$ more parameters.
All code, models, and datasets will be made publicly available.

\end{abstract}
\section{Introduction}
\label{sec: intro}
\vspace{-1.5pt}
Recent advances have highlighted the potential of rule-based Reinforcement Learning (RL) to elicit reasoning capabilities of Large Language Models (LLMs)~\citep{guo2025deepseek,yu2025dapo}. 
In particular, DeepSeek-R1~\citep{guo2025deepseek} exemplifies the success of applying the GRPO \citep{shao2024deepseekmath} reinforcement learning algorithm to incentive strong reasoning with long Chain-of-Thought (CoT) in LLMs. 
Beyond text-based domains, this paradigm has also shown promising results in Multimodal Large Language Models (MLLMs), with representative models including R1-OneVision~\citep{yang2025r1}, OpenVLThinker~\citep{deng2025openvlthinker}, and Video-R1~\citep{feng2025video}. The key of these methods is to utilize a rule-based function that yields accurate outcome reward signals for RL training~\citep{guo2025deepseek,MMR1-Math2025,deng2025openvlthinker}.

\begin{figure}
  \centering
  \includegraphics[width=0.95\textwidth]{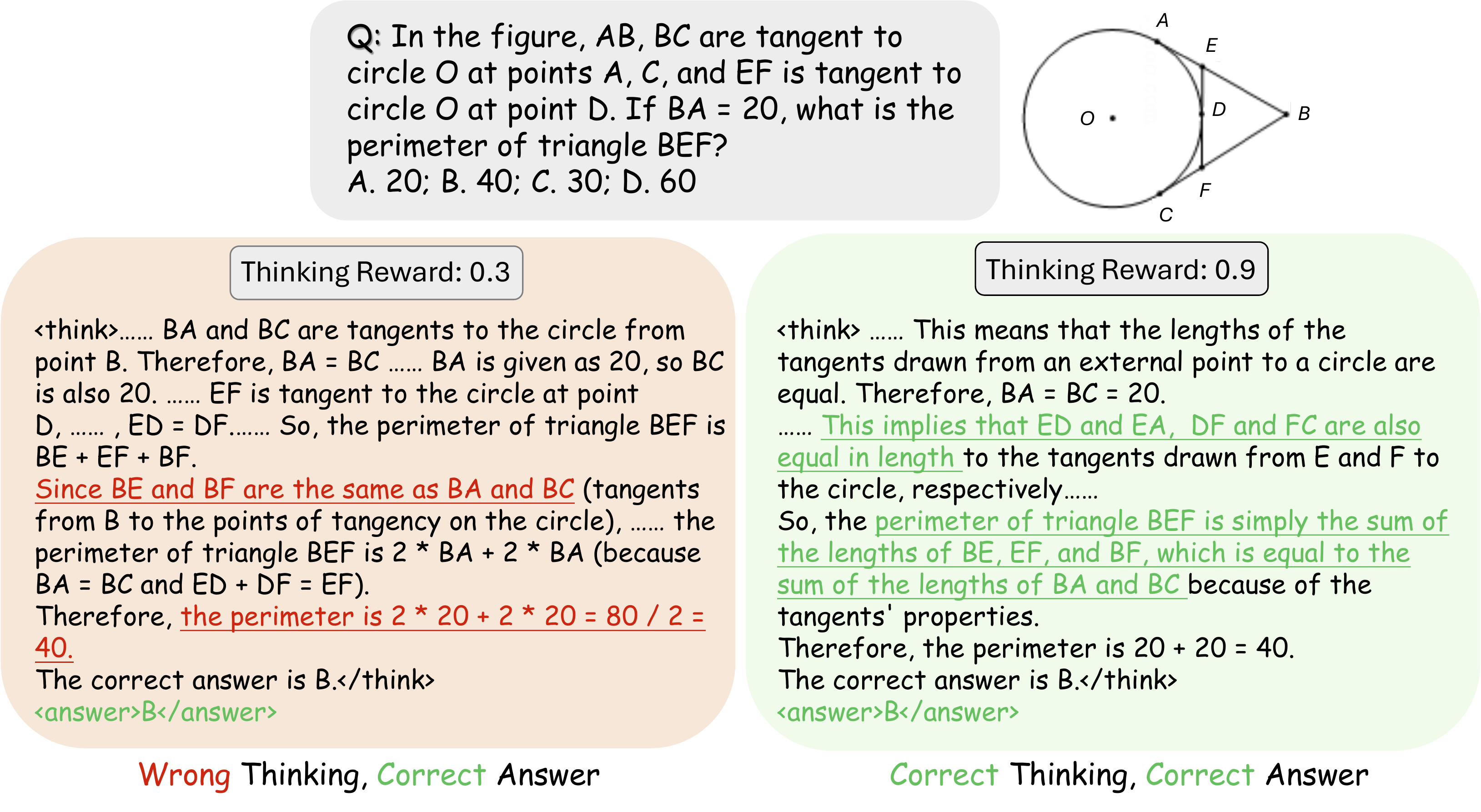}
  \caption{Examples of model responses and their corresponding thinking rewards. }
  \label{fig: demo}
\end{figure}

However, solely relying on the outcome reward usually fails to ensure the quality of the thinking process, which is critical for developing models with generalizable reasoning ability~\citep{lightman2023let}. For example, models may produce correct answers through flawed thinking trajectories, as illustrated in Figure~\ref{fig: demo}, rather than through systematic deduction. 
During GRPO training~\citep{shao2024deepseekmath}, the rule-based outcome reward will equally encourage these responses with correct answers, regardless of whether the underlying thinking process is sound or flawed.
Therefore, the model may adopt sub-optimal or even wrong reasoning strategies that generalize poorly, leading to inferior performance.
This gives rise to one intuitive thought: \textit{Can we incorporate a reward for the thinking process during GRPO training to explicitly guide correct reasoning?}

To explore this question, we propose \textbf{SophiaVL-R1}, an MLLM that enhances reasoning by integrating model-generated thinking rewards with rule-based outcome rewards in RL training. 
Given that typical process reward models (PRMs) impose rigid step-wise constraints on reasoning and can be overly exploited (\textit{e.g.}, generating meaningless or repetitive steps), we measure the quality of the thinking process at a \textit{holistic} level rather than at the \textit{step} level.
Specifically, we introduce a thinking reward model trained on annotated reasoning responses collected from GRPO training trajectories.
This model evaluates intermediate reasoning quality based on criteria such as logical soundness, consistency across steps, and redundancy in the thinking process. By doing so, we provide reward signals that help the reasoning model distinguish between sound and flawed thinking processes.

Moreover, considering that the model-generated thinking rewards may be unreliable for certain cases~\citep{ye2024justice,li2025preference}, we propose the \textbf{Trust-GRPO} training algorithm to reduce the risks of reward hacking~\citep{skalse2022defining}. The core idea of Trust-GRPO is to add a trustworthiness weight to the thinking reward, which evaluates the reliability of the rewards across a group of responses to a given question. 
This weight is determined by comparing the thinking rewards of responses that produce correct answers with those that yield incorrect answers for the same question.
A lower trustworthiness weight is assigned when high thinking rewards are abnormally given to reasoning processes that lead to incorrect answers, indicating that the reward signal may be unreliable. 
Unlike previous uncertainty estimation methods such as  MC Dropout~\citep{gal2016dropout}, which usually require multiple samplings for a single response—an approach that can be computationally prohibitive for MLLMs—our method introduces no additional cost by leveraging information from the response group within GRPO. 
Furthermore, an annealing schedule is introduced to gradually reduce the influence of the thinking reward throughout training, allowing the model to increasingly rely on the more reliable and accurate rule-based outcome reward in later stages.
In short, our proposed Trust-GRPO enables the model to receive thinking process rewards in a reliable manner, thereby guiding the exploration of favorable and generalizable reasoning strategies.
 
In summary, our contributions are as follows:
\begin{itemize}
    \item We propose a thinking reward model that evaluates reasoning quality from various dimensions at a holistic level, enabling the model to distinguish between sound and flawed reasoning processes during rule-based RL training.
    \item We introduce the Trust-GRPO algorithm, which assigns a trustworthiness weight to thinking rewards based on their reliability. This method guides the model to explore favorable reasoning policies in a trustworthy manner without extra computational overhead.
    \item In the experiments, SophiaVL-R1-7B consistently outperforms existing MLLMs on diverse benchmarks (\textit{e.g.}, MathVista, MMMU), highlighting its strong reasoning and generalization abilities.
    Notably, our SophiaVL-R1-7B outperforms LLaVA-OneVision-72B with $10\times$ more parameters on most benchmarks.
\end{itemize}

\section{Related Work}
\vspace{-1.5pt}
\subsection{Process Reward Models}
\vspace{-1.5pt}

Reward models (RMs) play a crucial role in guiding and shaping the behavior of models ~\citep{ouyang2022training,zhong2025comprehensive}.
Several studies~\citep{lightman2023let,yuan2024free,wang2025visualprm,zhang2025lessons} demonstrate that process supervision—providing feedback at intermediate reasoning steps—has the potential to enhance reasoning capabilities.
For example, \citet{lightman2023let} introduce powerful Process Reward Models (PRMs) with step-wise rewards, which have been applied to mathematical reasoning~\citep{lightman2023let,wang2023math}.
ReST-MCTS*~\citep{zhang2024rest} integrates process supervision and Monte Carlo Tree Search (MCTS) to generate per-step process rewards, enabling efficient self-training of both policy and reward models without manual annotation.
Beyond the text-based domain, VisualPRM~\citep{wang2025visualprm} extends PRMs to the multimodal domain, achieving significant improvements in the reasoning performance of various MLLMs.
Despite these advances, PRMs still face two major challenges: (1) imposing rigid step-wise constraints requires the model to strictly follow predefined reasoning steps, which can limit flexibility and generalization—particularly in general tasks~\citep{guo2025deepseek,cui2025process}; and (2) evaluating the correctness of individual steps is inherently challenging~\citep{zhong2025comprehensive}, which may lead models to exploit the reward by repeating valid steps or inserting meaningless ones without making real progress. Therefore, in contrast to prior approaches, we aim to develop a thinking reward model that evaluates reasoning quality from multiple dimensions at a holistic level.
\vspace{-1.5pt}
\subsection{Multimodal Large Language Model Reasoning}
\vspace{-1.5pt}

The field of multimodal large language model reasoning aims to build human-like models capable of handling complex tasks that require understanding and reasoning across multiple modalities~\citep{li2025perception}. 
Earlier methods typically depend on fine-grained step-level supervision or learned reward models to guide the reasoning process \citep{yao2024mulberry,wang2025visualprm,zang2025internlm}.
In contrast, DeepSeek-R1~\citep{guo2025deepseek} demonstrates that reinforcement learning with a rule-based reward model can effectively incentivize strong reasoning abilities without dense supervision.
Following the R1 paradigm, several efforts have explored enhancing MLLM reasoning through rule-based reinforcement learning~\citep{lai2025med,feng2025video,shen2025vlm,xia2025gui,wang2025simplear}. 
R1-OneVision~\citep{yang2025r1} introduces a cross-modal reasoning pipeline and adopts a supervised fine-tuning followed by RL strategy to strengthen reasoning capabilities. Curr-ReFT~\citep{wu2025boosting} introduces a curriculum-based reinforcement learning paradigm for small-scale MLLMs, combining difficulty-aware rewards and rejection sampling to boost generalization. 
Video-R1~\citep{feng2025video} proposes T-GRPO algorithm to explicitly encourage temporal reasoning in video.
Despite their success on multimodal tasks, these approaches rely exclusively on outcome rewards, which often overlook the quality of intermediate reasoning steps. 

\section{Method}
\label{sec: method}

\vspace{-1.5pt}

\subsection{Dataset Composition}
\label{sec:method.dataset_composition}
\begin{figure}
  \centering
  \includegraphics[width=\textwidth]{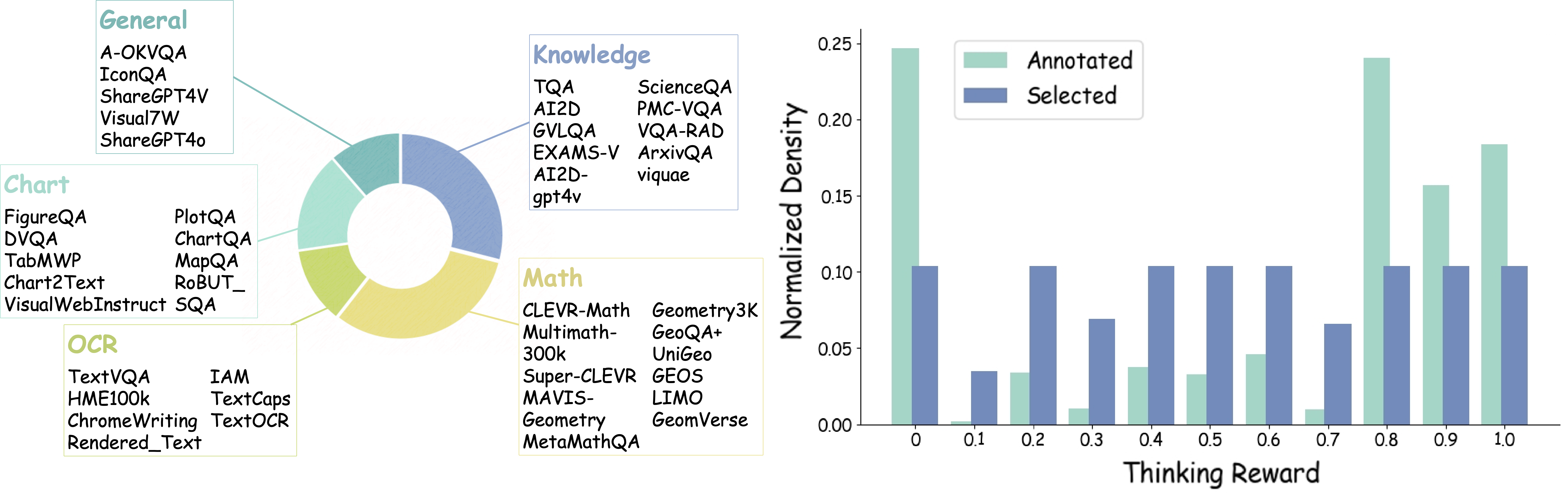}
  \caption{\textbf{Left:} Composition of our SophiaVL-R1-130k dataset from public sources. \textbf{Right:} Distribution of the SophiaVL-R1-Thinking-156k dataset used to train the thinking reward model.}
  \label{fig:dataset}
\end{figure}

\vspace{-1.5pt}
We curate a dataset SophiaVL-R1-130k, comprising 130k examples to support the training of thinking reward model (Section~\ref{sec:method.think_reward_model}) and SophiaVL-R1 (Section~\ref{sec:method.hyart}).
To overcome the scarcity of high-quality multimodal reasoning data and ensure robust model performance across a wide range of tasks, we aggregate samples from a combination of text-only and multimodal datasets, all of which are publicly available. The dataset contains both reasoning-specific tasks and general vision-language understanding tasks.
We organize the data into five categories, covering diverse reasoning scenarios, as illustrated in Figure~\ref{fig:dataset} (left).

\vspace{-1.5pt}
\subsection{Thinking Reward}
\label{sec:method.think_reward_model}
\vspace{-1.5pt}
To assess fine-grained reasoning quality of MLLMs' thinking process, we develop a thinking reward model that assigns a score between 0 and 1 based solely on the quality of intermediate reasoning, regardless of whether the final answer is correct.

To construct the dataset used for training the thinking reward model, we collected 470,331 (\texttt{question, response}) pairs output by Qwen2.5-VL-7B-Instruct \citep{bai2025qwen2} during the GRPO training
on the SophiaVL-R1-130k dataset. 
These data contain both favorable and flawed reasoning patterns occurred in the training. Then, each response is scored by the advanced MLLM, Qwen2.5-VL-72B-Instruct~\citep{bai2025qwen2}, using the prompt in Appendix~\ref{suppl: prompt}. This results in 470,331 (\texttt{question, response, thinking reward}) tuples.
The evaluation is based on five dimensions, which are identified from error patterns observed during GRPO training: Logical Soundness, Correct Reasoning, Error Identification, Language Consistency, and Redundancy. Detailed examples of error patterns are provided in Appendix~\ref{suppl: error pattern in grpo}.

To ensure the quality of labels and maintain a balanced distribution across different reward levels, we apply manually designed rule-based filtering to remove noisy samples and perform uniform sampling to preserve distribution balance. This process results in 156,703 high-quality annotated samples.
with 5,000 to 15,000 samples per interval. Each reward interval corresponds to a discrete range (\textit{e.g.}, [0.0–0.1), [0.1–0.2), ..., [0.9–1.0]). The distribution of the full (\textit{Annotated}) and balanced (\textit{Selected}) datasets is shown in Figure~\ref{fig:dataset} (right). 
We denote the resulting dataset as SophiaVL-R1-Thinking-156k, with its detailed composition reported in Appendix~\ref{suppl:dataset}.

The thinking reward model, initialized with Qwen2.5-VL-3B-Instruct~\citep{bai2025qwen2}, is then trained on this dataset using SFT, where the model is required to output a thinking reward given a question and its corresponding thinking process.
Through this training, the thinking reward model learns to identify diverse reasoning errors and assign appropriate rewards accordingly, thereby playing a crucial role in GRPO training by providing feedback on reasoning quality.

\vspace{-1.5pt}
\subsection{Rule-based Outcome Reward}
\label{sec: method.ttrm.or}
\vspace{-1.5pt}

Following DeepSeek-R1~\citep{guo2025deepseek}, we construct rule-based outcome reward functions to generate reward signals. Specifically, we design task-specific functions that assess model outputs by comparing them with ground-truth answers. 
Tasks are categorized based on their output formats:
(1)\textbf{Numerical}: A binary reward is assigned based on an exact match between the predicted and ground-truth values;
(2) \textbf{Multiple Choice}: The reward is defined based on whether the model's output matches the ground-truth choice;
(3) \textbf{OCR}: The reward is computed as the negative Word Error Rate (WER), penalizing transcription inaccuracies;
(4) \textbf{Free-form Text}: The reward is calculated as the average of ROUGE-1, ROUGE-2, and ROUGE-L scores, measuring n-gram and sequence-level similarity~\citep{feng2025video}.

\vspace{-1.5pt}
\subsection{Trustworthy Group Relative Policy Optimization (Trust-GRPO)}
\label{sec:method.hyart}
\vspace{-1.5pt}

As discussed earlier, integrating the thinking reward into GRPO training could help the model distinguish between favorable and flawed reasoning process.  Nevertheless, a direct application may result in reward hacking, given that model-generated rewards are not always trustworthy. To deal with this challenge, we introduce the Trust-GRPO algorithm, as illustrated in Figure \ref{fig: overview}.

Trust-GRPO optimizes the policy using a combination of two reward types: (1) thinking reward $R^t$ (Section~\ref{sec:method.think_reward_model}) that assigns a score between 0 and 1 based on holistic reasoning quality, and (2) outcome reward $R^o$ (Section~\ref{sec: method.ttrm.or}), derived from rule-based evaluation of outcome answer correctness. To reduce the risk of reward hacking, a trustworthiness weight $\gamma$ is included to determine the influence of thinking reward $R^t$.

The trustworthiness is computed by contrasting the thinking reward $R^t$ assigned to responses that arrive at correct answers with those leading to incorrect ones. 
When higher thinking rewards are abnormally associated with incorrect reasoning, $\gamma$ will be lower, indicating the potential unreliability in the reward signal. Next, we will introduce how to derive it.

First, responses $o_i$ to a question $q$ are grouped into correct answer group $G_{\text{correct}}$ and wrong answer group $G_{\text{wrong}}$ based on their outcome rewards. Then, we calculate the average thinking reward in $G_{\text{correct}}$ and $G_{\text{wrong}}$ as follows:
\begin{align}
\mu_c &= \frac{1}{|G_{\text{correct}}|} \sum_{i \in G_{\text{correct}}} R_i^t, \quad 
G_{\text{correct}} = \left\{ i \mid R_i^o \ge 0.5 \right\}, \\
\mu_w &= \frac{1}{|G_{\text{wrong}}|} \sum_{i \in G_{\text{wrong}}} R_i^t, \quad 
G_{\text{wrong}} = \left\{ i \mid R_i^o < 0.5 \right\},
\end{align}

where $\mu_c$ and $\mu_w$ denote the average thinking rewards in the correct answer group and the wrong answer group, respectively. $R_i^o$ denotes the outcome reward of response $i$. The trustworthiness weight $\gamma$ is defined as follows:
\vspace{-1pt}
\begin{figure}
  \centering
  \includegraphics[width=\textwidth]{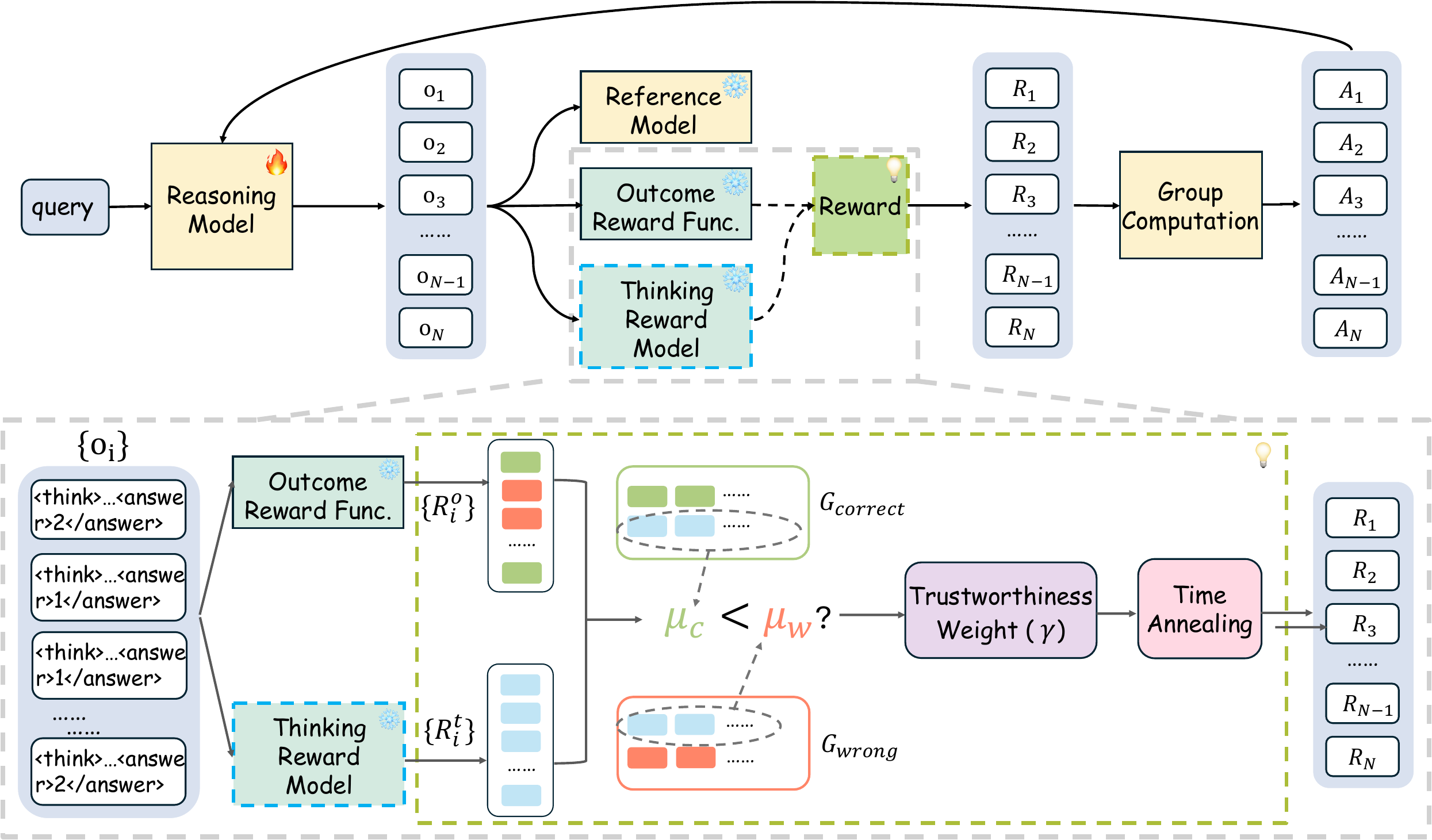}
  \caption{An illustration of our proposed Trust-GRPO.}
  \label{fig: overview}
\end{figure}

\begin{equation}
    \gamma = \begin{cases}
        1, & \mu_c\ge \mu_w \\
        e^{{\mu_c}-\mu_w}, & \mu_c < \mu_w
    \end{cases}.
\end{equation}

This comparison between $\mu_c$ and $\mu_w$ allows us to assess the alignment between thinking rewards and rule-based outcome rewards. A lower $\gamma$ indicates a discrepancy between $R^t$ and $R^o$, suggesting that the thinking reward may be unreliable for this response group and thus should be given reduced weight. $\gamma$ dynamically estimates the trustworthiness of thinking rewards for each question's response group without incurring additional computational overhead. 
This average reward–based design is chosen for its simplicity and efficiency, which are particularly important given the high cost of training and inference of MLLMs. 
Additional analyses of alternative design are provided in Appendix~\ref{suppl:trustworthiness}.

The $i$-th reward $R_i$ incorporating the thinking reward with trustworthiness weight is defined as:
\vspace{-1.5pt}
\begin{equation}
    R_i = R_i^o + \gamma \alpha \cdot R_i^t,
    \label{eq:final_reward}
\end{equation}
where $\alpha$ is a hyperparameter that controls the impact of thinking reward.

We further introduce a time-based annealing strategy that gradually reduces the influence of thinking reward as training progresses. This encourages the reasoning model to rely increasingly on the more accurate rule-based outcome reward in later steps. Combining both components, the final reward $R_i$ is defined as: 
\begin{equation}
    R_i = R_i^o + \gamma \alpha e^{-\frac{\texttt{steps}}{T}} \cdot R_i^t,
    \label{eq:adaptive_reward}
\end{equation}

where \texttt{steps} denotes the current global training step and $T$ is the total number of training steps, controlling the decay rate of thinking reward's influence over time. 
Additional analysis of the time-based annealing strategy design is provided in Appendix~\ref{suppl:time_based}.

\begin{figure}
  \centering
  \includegraphics[width=\textwidth]{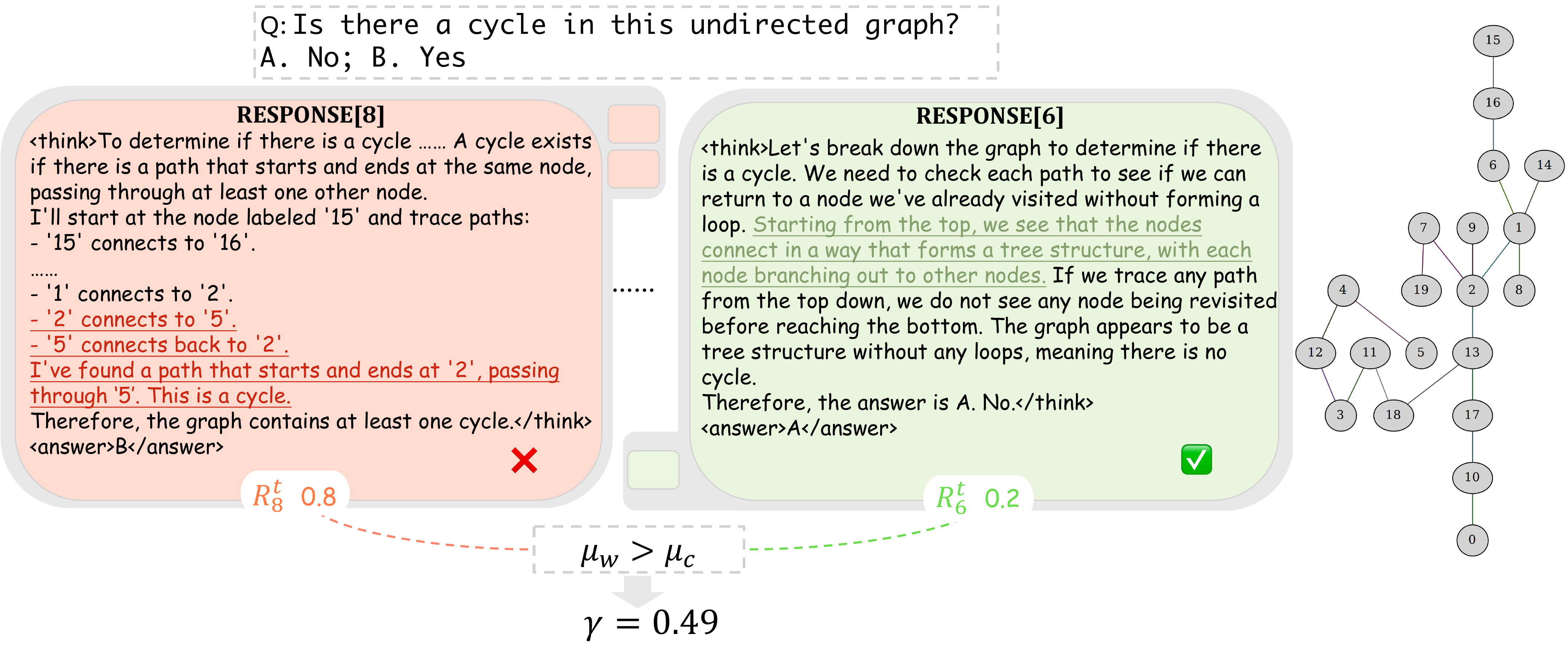}
  \caption{Example of trustworthiness weight $\gamma$. Incorrect responses (red) receive higher average thinking rewards than correct ones (green), indicating misalignment between $R^t$ and $R^o$ and the need for a trustworthiness-aware adjustment.}
  \label{fig: gamma}
  \vspace{-0.2in}
\end{figure}

The advantage ${A}_{i}$ is computed using rewards of each response group:
\begin{equation}
{A}_{i} =  \frac{R_i - \text{mean}(\{R_1,R_2,\cdots,R_N\})}{\text{std}(\{R_1,R_2,\cdots,R_N\})},
\end{equation}
\vspace{-0.5pt}
Following DeepSeek-R1~\citep{guo2025deepseek}, given a question $q$, GRPO samples responses ${o_1,  \ldots, o_N}$ from the old policy $\pi_{\text{old}}$, and updates the policy $\pi_\theta$ by maximizing the following objective:
\begin{align}
    \mathcal{J}&_{GRPO}(\theta)  = \mathbb{E}\left[q \sim P(Q),\ \{o_i\}_{i=1}^N \sim \pi_{\text{old}}(O|q)\right] \nonumber \\
    & \frac{1}{N} \sum_{t=1}^N \left( \min\left(
        \frac{\pi_{\theta}(o_{i}|q)}
             {\pi_{\text{old}}(o_{i}|q)} {A}_{i}, \right. \left. \text{clip}\left(
            \frac{\pi_{\theta}(o_{i}|q)}
                 {\pi_{\text{old}}(o_{i}|q)},
            1 - \epsilon,\ 1 + \epsilon
        \right) {A}_{i} \right) 
        - \beta\, \mathbb{D}_{\text{KL}}[\pi_{\theta} \| \pi_{\text{ref}}] \right).
\end{align}
\vspace{-0.5pt}

By contrasting the thinking rewards of correct and incorrect responses, Trust-GRPO improves the reliability of reward signals, thereby encouraging more generalizable reasoning behavior.

Figure~\ref{fig: gamma} illustrates a case where the trustworthiness weight $\gamma$ helps identify potentially unreliable thinking rewards. Responses with incorrect answers are shown in red and those with correct answers in green. Despite being incorrect, the red group receives a higher average thinking reward, indicating a misalignment between $R^t$ and $R^o$. 
This implies that the thinking reward has potential risk of unreliability, thus should be assigned less weight. More examples can be found in Appendix~\ref{suppl: case gamma}.

\section{Experiment}
\vspace{-1.5pt}
\subsection{Experiment Settings}
\label{sec: exp setting}
\vspace{-1.5pt}
\textbf{Benchmarks.}
We evaluate our model on both multimodal mathematical reasoning and general multimodal reasoning benchmarks. For mathematical reasoning, we report detailed results on MathVista~\citep{lu2023mathvista} and MathVerse~\citep{zhang2024mathverse}. For general multimodal capabilities, we conduct evaluations on MMMU~\citep{yue2024mmmu}, MME~\citep{liang2024survey}, MMStar~\citep{chen2024we}, ChartQA~\citep{masry2022chartqa}, and MMBench~\citep{xu2023mmbench}.

\vspace{-1.5pt}
\paragraph{Implementation Details.}
The thinking reward model is initialized from Qwen2.5-VL-3B-Instruct and trained for 2 epochs with SFT on SophiaVL-R1-Thinking-156k using 4 NVIDIA A800 80GB GPUs.
The reasoning model is initialized from Qwen2.5-VL-7B-Instruct and trained on SophiaVL-R1-130k  with the Trust-GRPO algorithm. RL training is performed for 1,500 steps using a VeRL~\citep{zheng2025easyr1,sheng2024hybridflow}-based implementation on 8 NVIDIA A800 80GB GPUs. Hyperparameters for RL training are provided in Appendix~\ref{suppl:train-details}.
For evaluation, we use default prompts to generate responses. Additional evaluation details are given in Appendix~\ref{suppl:eval-details}.

\begin{table}
\centering
\small
\caption{Comparison of models on \textbf{MathVista} and \textbf{MathVerse}. The best is \textbf{bold}, and the runner-up is \underline{underline}.
\vspace{-1pt}
\footnotesize
\textsuperscript{1}Scientific Reasoning, 
\textsuperscript{2}Textbook Question Answering, 
\textsuperscript{3}Arithmetic Reasoning, 
\textsuperscript{4}Math Word Problem, 
\textsuperscript{5}Logical Reasoning, 
\textsuperscript{6}Vision Intensive, 
\textsuperscript{7}Vision Only, 
\textsuperscript{8}Vision Dominant, 
\textsuperscript{9}Text Dominant, 
\textsuperscript{10}Text Lite.}
\resizebox{\linewidth}{!}{%
    \setlength{\tabcolsep}{1.2mm}
     \renewcommand\arraystretch{1.4}
\begin{tabular}{l|*{6}{c}|*{6}{c}}
\toprule
\multirow{2}{*}{Model} & \multicolumn{6}{c|}{MathVista} & \multicolumn{6}{c}{MathVerse} \\
\cmidrule{2-13}
 & AVG & SCI\textsuperscript{1} & TQA\textsuperscript{2} & ARI\textsuperscript{3} & MWP\textsuperscript{4} & LOG\textsuperscript{5} & AVG & VI\textsuperscript{6} & VO\textsuperscript{7} & VD\textsuperscript{8} & TD\textsuperscript{9} & TL\textsuperscript{10} \\
\midrule
\multicolumn{13}{l}{\textit{General MLLMs}} \\
LLaVA-OneVision-7B~\citep{li2024llava} & 63.2  & 65.6& 60.8& 57.8& 69.4& 21.6 &26.2 & -& -& -& -& -\\
LLaVA-OneVision-72B~\citep{li2024llava} &  68.4 & 63.1& 65.8& 60.1 & 73.7& \underline{27.1}& 27.2& -& -& -& -& - \\
Cambrian-1-34B~\citep{tong2024cambrian}&50.9  & 53.3& 55.1& 45.6& 51.6 & 16.2& -  & -& -& -& -& -\\
GPT-4V & 51.8  & 63.1& 65.8& 51.8& 57.5& 21.6& 32.8 & -& -& -& -& -\\
\midrule
\multicolumn{13}{l}{\textit{Open-Source Math MLLMs}} \\
Math-LLaVA-13B~\citep{shi2024math} & 46.6 & 49.2 & 51.3 & 40.2 & 56.5 & 16.2 & 22.9 & 24.5 & 16.1 & 21.7 & 27.3 & 24.9 \\
Math-PUMA-Qwen2VL-7B~\citep{zhuang2025math} & 47.9 & 42.6 & 46.2 & 46.2 & 68.3 & 21.6 & 33.6 & 33.4 & 26.0 & 31.6 & 42.1 & 35.0 \\
Multimath-7B~\citep{peng2024multimath}& 50.0 & - & 50.0 & - & 61.8 & - & 26.9 & 28.1 & 15.0 & 25.9 & 34.8 & 30.8 \\
URSA-8B~\citep{luo2025ursa} & 59.8 & 58.2 & 63.9 & 53.5 & 75.3 & 21.6 & 45.7 & \textbf{46.4} & 34.6 & \underline{43.9} & 55.3 & \underline{48.3} \\
\midrule
\multicolumn{7}{l}{\textit{Open-Source Reasoning MLLMs}} \\
Curr-ReFT-7B~\citep{deng2025boosting}&  64.5 & -& -& -& -& -&  - & -& - & -& -& -\\ 
R1-OneVision-7B~\citep{yang2025r1} &64.1 & 61.5& 62.0 & 56.1 & 64.5 & 16.2 & \underline{46.4} & -& 40.0& -& -& - \\ 
InternVL2.5-8B-VisualPRM~\citep{wang2025visualprm} & 68.5 & 61.5& 53.9&45.9 & 66.8& 21.2 & 30.7 & 28.9 & 35.8 & 27.3& 31.7& 29.7 \\ 
\midrule
Qwen2.5-VL-7B-Instruct~\citep{bai2025qwen2} & 67.5 & 65.6 & 67.7 & 57.5 & 69.4 & 27.0 & 44.0 & 41.1 & 41.0 & 38.7 & 55.2 & 44.0 \\
\quad +GRPO & \underline{69.9} & 68.0 & 69.6 & \underline{61.2} & \underline{75.8} & 24.3 & 45.3 & 43.0 & \underline{41.0} & 41.1 & \underline{56.0} & 45.6 \\
\quad +SFT+GRPO & 66.8 & \textbf{72.1} & \underline{73.4}& 59.8& 69.9 & 21.6 & 43.1 & 42.5 & 37.1 & 37.3 & 52.2 & 46.3 \\
\rowcolor{mypurple} {\MODEL}-7B & \textbf{71.3} & \underline{70.5} & \textbf{73.4} & \textbf{62.6} & \textbf{76.9} & \textbf{35.1} & \textbf{48.8} & \underline{45.4} & \textbf{43.9} & \textbf{45.1} & \textbf{58.5} & \textbf{51.3} \\
\bottomrule
\end{tabular}
}
\label{tab:math results}
\vspace{-0.1in}
\end{table}
\vspace{-1.5pt}
\subsection{Main Results}
\vspace{-1.5pt}

\paragraph{Performance on Math Reasoning Benchmarks.}
\label{sec: exp-math-reasoning}
As shown in Table~\ref{tab:math results}, {\MODEL}-7B achieves competitive performance on mathematical reasoning benchmarks. On the MathVista benchmark, it attains an accuracy of 71.3\%, surpassing both Qwen2.5-VL-7B-Instruct models trained with GRPO and SFT+GRPO, and also outperforming the LLaVA-OneVision-72B model.
Compared to the model trained by VisualPRM~\citep{wang2025visualprm}, our model achieves significantly better performance, with an 18.1-point improvement on MathVerse (48.8 vs. 30.7), and consistently outperforms it across all sub-tasks. These results indicate that, compared to PRM-based method, our Trust-GRPO may serve as a more effective approach for providing reward signals, better guiding the model toward improved reasoning ability.

\vspace{-1.5pt}
\textbf{Performance on General Benchmarks.}
Many task-specific reasoning models, such as those optimized for mathematical problem-solving or other specialized tasks, excel within their respective domains but often struggle to maintain strong performance on general multimodal benchmarks (\textit{e.g.}, URSA-8B).
Different from them, {\MODEL}-7B demonstrates consistently strong performance across widely recognized general ability benchmarks, as shown in Table~\ref{tab: general results}, highlighting its superior generalization capability. For example, on the widely used MMMU benchmark for multi-discipline reasoning, SophiaVL-R1-7B outperforms LLaVA-OneVision-72B by 4.5 points.

\vspace{-1.5pt}
\subsection{Performance of Thinking Reward Model}
\vspace{-1.5pt}
To further evaluate the capability of our thinking reward model, we conduct experiments on VLRewardBench~\citep{li2025vl}, a benchmark designed to assess multimodal reward models. 

As shown in Table~\ref{tab:vlrewardbench}, our 3B thinking reward model achieves higher performance despite having significantly fewer parameters. In particular, it demonstrates strong performance in detecting Hallucination, indicating that it effectively distinguishes reliable from unreliable responses.


\begin{table}
\centering
\small
\caption{Comparison on general ability benchmarks. The best is \textbf{bold}, and the runner-up is \underline{underline}.}
\vspace{-1.5pt}
\resizebox{\linewidth}{!}{%
    \setlength{\tabcolsep}{1.2mm}
     \renewcommand\arraystretch{1.4}
\begin{tabular}{llcccccc}
\toprule
\textbf{Model} & \textbf{MMMU} & \textbf{MME} & \textbf{ChartQA} & \textbf{MMBench} & \textbf{MMStar} \\
\midrule
\multicolumn{6}{l}{\textit{General MLLMs}} \\
LLaVA-OneVision-7B~\citep{li2024llava} & 48.8 & 1998.0 & 80.0 & - & 61.7 \\
LLaVA-OneVision-72B~\citep{li2024llava} & 56.8 & 2261.0 & 83.7 & - & \underline{66.1}\\
Cambrian-1-34B~\citep{tong2024cambrian} & 49.7 &1689.3 & 75.6 & 81.4 & 54.2 \\
GPT-4V & 56.8&1926.0 & 78.5 & 75.0 & 57.1 \\
\midrule
\multicolumn{6}{l}{\textit{Open-Source Math MLLMs}} \\
URSA-8B~\citep{luo2025ursa} & 43.1 & 1605.7 & 44.4 & 55.5 & 42.3 \\
\midrule
\multicolumn{6}{l}{\textit{Open-Source Reasoning MLLMs}} \\
Curr-ReFT-7B~\citep{deng2025boosting} & - & -  &- & 79.0 & -\\
R1-Onevision-7B~\citep{yang2025r1} & 51.6  & 2223.3 & - & 75.6 &59.1\\
InternVL2.5-8B-VisualPRM~\citep{wang2025visualprm} & 56.2 & - & 60.8 &83.5 & 63.4\\
\midrule
Qwen2.5-VL-7B-Instruct~\citep{bai2025qwen2} & 58.7  & 2306.0 & 86.3 & 83.3  & 64.3\\
\quad +GRPO  & 58.0 & 2298.2 & 87.2& 83.4 & 65.6 \\
\quad +SFT+GRPO & \underline{59.1} & \underline{2344.1} & \textbf{89.2} & \underline{84.6}& 64.7\\
\rowcolor{mypurple} {\MODEL}-7B & \textbf{61.3} & \textbf{2403.8}& \underline{88.5} & \textbf{85.4} & \textbf{66.7}\\
\bottomrule
\end{tabular}
}
\label{tab: general results}
\vspace{-0.1in}
\end{table}

\begin{table}
\centering
\caption{Performance of reward models on VLRewardBench. }
\vspace{-1.5pt}
\resizebox{\linewidth}{!}{%
    \setlength{\tabcolsep}{1.2mm}
     \renewcommand\arraystretch{1.4}
\begin{tabular}{lccccc}
\hline
\textbf{Model} & \textbf{General} & \textbf{Hallucination} & \textbf{Reasoning} & \textbf{Overall Accuracy} & \textbf{Macro Accuracy} \\
\hline
Qwen2.5-VL-3B-Instruct & 34.4 & 42.1 & 51.5 &43.1 & 43.0\\
GPT-4o-mini                    & 41.7    & 34.5          & 58.2      & 41.5             & 44.8           \\
Qwen2-VL-72B                   & 38.1    & 32.8          & 58.0      & 39.5             & 43.0           \\
\rowcolor{mypurple} Our Thinking Reward Model (3B) & 45.4    & 46.8          & 54.4      & 48.6             & 48.9           \\
\hline
\end{tabular}}
\vspace{-0.15in}
\label{tab:vlrewardbench}
\end{table}
\vspace{-1.5pt}
\section{Ablation Study}
\label{sec: ablation}
\vspace{-1.5pt}

We conduct ablation studies to examine the contributions of key components in our method. Specifically, we evaluate three variants of our SophiaVL-R1:
\vspace{-2pt}
\begin{itemize}
\item \textbf{SophiaVL-R1-wo-trained-TRM}: replacing the trained thinking reward model with an untrained Qwen2.5-VL-3B-Instruct model.
\item \textbf{SophiaVL-R1-wo-trust-and-annealing}: removing both the trustworthiness weighting and the annealing strategy from Trust-GRPO.
\item \textbf{SophiaVL-R1-wo-trust}: removing only the trustworthiness weight while retaining the time-based annealing schedule.
\end{itemize}

\vspace{-2pt}

Besides, we also include \textbf{Qwen2.5-VL-7B+GRPO} as a baseline, which directly uses GRPO for training Qwen2.5-VL-7B-Instruct. The results are summarized in Table~\ref{table: ablation}.

\vspace{-2pt}

\paragraph{Effect of the Thinking Reward Model.}
SophiaVL-R1-wo-trained-TRM consistently underperformances SophiaVL-R1. 
This highlights the effectiveness of our training pipeline and the SophiaVL-R1-Thinking-156k dataset in improving thinking reward model’s ability to provide accurate and informative reward signals for reasoning optimization.
What's more, SophiaVL-R1-wo-trained-TRM performs comparably to the Qwen2.5-VL-7B+GRPO. This indicates that an untrained reward model provides limited guidance. In contrast, our trained thinking reward model substantially improves the model performance, which highlights its importance in our method.
\vspace{-1.5pt}
\begin{table}
\centering
\caption{Ablation Study.}
\vspace{-2pt}
\small
    \resizebox{\linewidth}{!}{%
    \setlength{\tabcolsep}{1.2mm}
     \renewcommand\arraystretch{1.4}
\begin{tabular}{llccccccc}
\toprule
\textbf{Model} & \textbf{MathVista} & \textbf{MathVerse} & \textbf{MMMU}  & \textbf{MME} & \textbf{ChartQA} & \textbf{MMBench} & \textbf{MMStar}\\
\midrule
Qwen2.5-VL-7B+GRPO & 69.9 & 45.3 & 58.0 & 2298.2 & 87.2 & 83.4 & 65.6\\
{\MODEL}-wo-trained-TRM & 68.4 & \underline{47.9} & 57.0 & 2347.1 &87.7 &\underline{84.0} &\underline{65.7} \\
{\MODEL}-wo-trust-and-annealing &67.4 & 46.3& 56.7& \underline{2366.8} & 86.3&  82.6&65.0 \\
{\MODEL}-wo-trust & \underline{70.2}& 47.8 & \underline{60.0} & 2363.3 & \underline{87.8} & 83.7&  65.2 \\
\rowcolor{mypurple} {\MODEL} & \textbf{71.3} & \textbf{48.8} & \textbf{61.3} & \textbf{2403.8} & \textbf{88.5} & \textbf{85.4} & \textbf{66.7} \\
\bottomrule
\end{tabular}
}
\vspace{-0.1in}
\label{table: ablation}
\end{table}

\vspace{-2pt}
\paragraph{Effect of the Trustworthiness Weight $\gamma$.} 
We observe a performance drop across all benchmarks in SophiaVL-R1-wo-trust when the trustworthiness weight is removed, compared to the full SophiaVL-R1 model. 
This demonstrates the effectiveness of trustworthiness weighting, which allows the model to receive thinking process rewards in a more reliable manner. 
\vspace{-2pt}

\paragraph{Effect of the Time-based Annealing Strategy.}
To assess the effect of time-based annealing, we compare SophiaVL-R1-wo-trust-and-annealing with SophiaVL-R1-wo-trust. SophiaVL-R1-wo-trust-and-annealing generally performs worse on most benchmarks.
The performance drop may be due to the over-exploitation of the thinking reward, where potentially unreliable signals could interfere with the optimization of the reasoning policy.
This suggests that gradually reducing the influence of the thinking reward by our proposed annealing strategy is beneficial, as it encourages reliance on the more reliable rule-based outcome reward in later training stages.
\vspace{-2pt}

\begin{figure}
    \centering
    \includegraphics[width=0.95\linewidth]{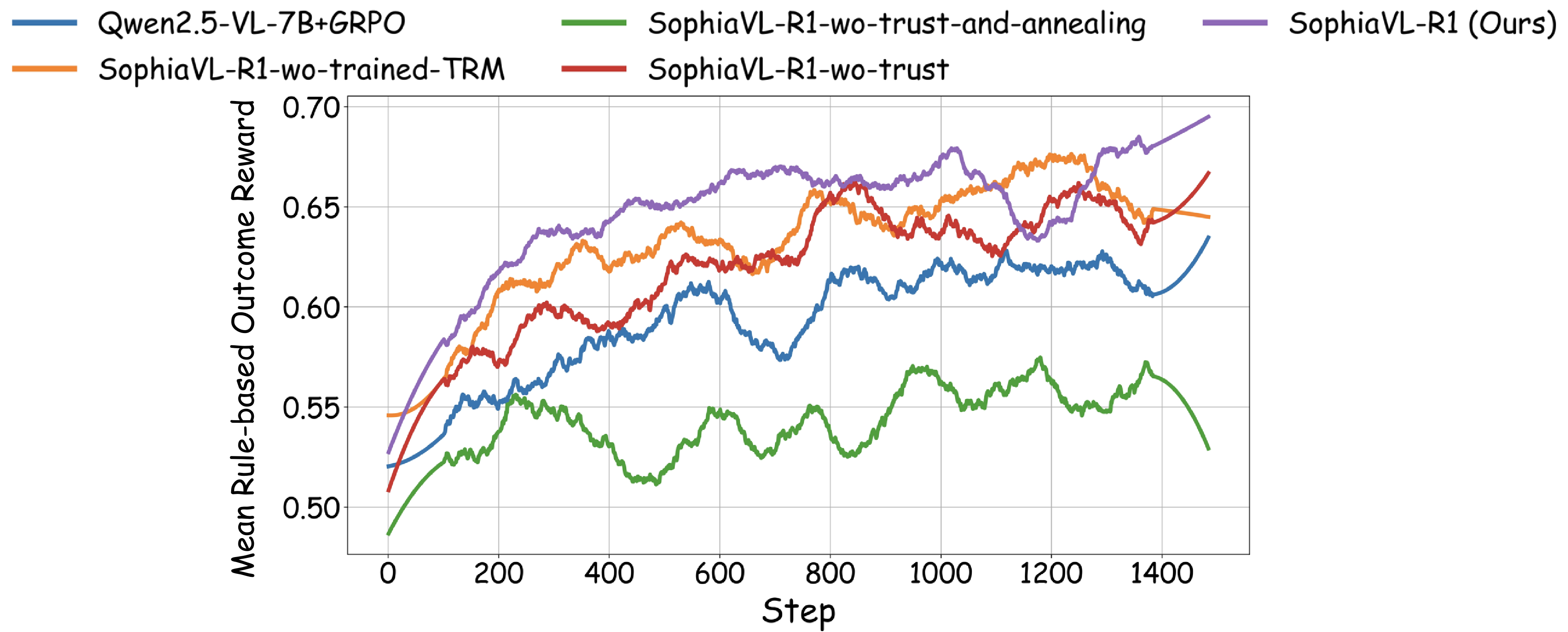}
    \vspace{-0.1in}
    \caption{Training curves of mean rule-based outcome reward across different methods.}
    \label{fig:curve}
    \vspace{-0.1in}
\end{figure}

\paragraph{Training Curve Analysis.} Figure~\ref{fig:curve} shows the mean outcome reward per training step for each method. 
SophiaVL-R1 achieves the highest reward and demonstrates faster improvement during training. 
Besides, we notice that directly combining thinking and outcome rewards (SophiaVL-R1-wo-trust-and-annealing) performs worse in training, indicating the effectiveness and necessity of our trustworthiness weighting and time-based annealing strategy.
Overall, these results underscore the importance of both Trust-GRPO and the thinking reward model.
\vspace{-2pt}

\section{Conclusion}
\vspace{-1.5pt}
In this work, we propose SophiaVL-R1, a multimodal large language model trained using a novel Trust-GRPO algorithm that integrates model-generated thinking rewards with rule-based outcome rewards. To promote generalizable reasoning, we introduce a holistic-level thinking reward model that assesses the quality of reasoning processes. Furthermore, we mitigate the challenge of reward hacking by introducing a trustworthiness weighting mechanism together with a time-based annealing strategy. 
Experimental results across multiple benchmarks demonstrate that SophiaVL-R1 consistently outperforms existing MLLMs. Our findings highlight the value of thinking process supervision beyond final correctness and offer insights for future studies on developing reasoning models.

\section{Acknowledgment}

This work is partially supported by the National Natural Science Foundation of China (No. 62306261), HK RGC-Early Career Scheme (No. 24211525), ITSP Platform Project (No. ITS/600/24FP) and the SHIAE Grant (No. 8115074). 
This study is supported in part by the Centre for Perceptual and Interactive Intelligence, a CUHK-led InnoCentre under the InnoHK initiative of the Innovation and Technology Commission of the Hong Kong Special Administrative Region Government. This work is partially supported by Hong Kong RGC Strategic Topics Grant (No. STG1/E-403/24-N), and CUHK-CUHK(SZ)-GDST Joint Collaboration Fund (No. YSP26-4760949).
This work is also supported by Shanghai Artificial Intelligence Laboratory.

\bibliography{iclr2026_conference}
\bibliographystyle{iclr2026_conference}

\newpage
\appendix

\section{Prompt Used for Evaluating Thinking Process Quality}
\label{suppl: prompt}
\begin{table}[H]
  \centering
    \caption{Prompt for evaluating thinking process quality.}
  \resizebox{\linewidth}{!}{%
    \setlength{\tabcolsep}{1.2mm}
     \renewcommand\arraystretch{1.4}
  \begin{tabular}{lp{0.68\linewidth}}
    \toprule
    \textbf{Input} & \{Image\}, \{Question\} and \{Model Response\}\\
    \midrule
    \multicolumn{2}{p{0.98\linewidth}}{You are an expert reasoning evaluator. I will give you a multimodal question and an answer. Your goal is to judge a reward process and give a score between 0 and 1. You should focus on whether the reasoning process is good rather than whether the final answer is correct.}\\
    \multicolumn{2}{p{0.98\linewidth}}{\textbf{Evaluation Criteria:}} \\
    \quad 1. Logical Soundness & Does each step follow logically from the previous one? \\
    \quad 2. Correct Reasoning & Are the methods and steps used appropriate and valid? Are the facts and lemmas correctly stated and applied? \\
    \quad 3. Error Identification & Are there logical flaws, unsupported assumptions, or incorrect steps? \\
    \quad 4. Language Consistency & Is the reasoning process conducted in a single, consistent language without mixing different languages? \\
    \quad 5. Redundancy & Is the reasoning concise, avoiding repetition or irrelevant steps? \\
    \multicolumn{2}{p{0.98\linewidth}}{Provide a single score from \textbf{\{0, 0.1, 0.2, \dots, 1.0\}} based on the reasoning quality, where:} \\
    \quad - 0 & Completely flawed reasoning. \\
    \quad - 1 & Perfectly sound reasoning. \\
    \quad - Intermediate & Reflect partial correctness or minor errors (\textit{e.g.}, 0.3 for significant flaws, 0.7 for minor errors). \\
    \multicolumn{2}{p{0.98\linewidth}}{Be strict, reward the good process and punish the bad one. You should only output the score without any explanation.} \\
    \bottomrule
  \end{tabular} }
  \label{tab: trm prompt}
\end{table}

\section{Training Details}
\label{suppl:train-details}
\begin{table}[ht]
\centering
\small
\caption{Training hyperparameters.}
\begin{tabular}{lc}
\toprule
\textbf{Hyperparameter} & \textbf{Value} \\
\midrule
Group size & 8 \\
Batch size & 8 \\
KL divergence coefficient & 0.04 \\
Learning rate & $5 \times 10^{-7}$ \\
$\alpha$ & 0.3 \\
Total training steps & 1500 \\
\bottomrule
\end{tabular}
\label{tab:hyperparams}
\end{table}

\section{Error Patterns Observed in GRPO Training}
\label{suppl: error pattern in grpo}

\begin{figure}
    \centering\includegraphics[width=\linewidth]{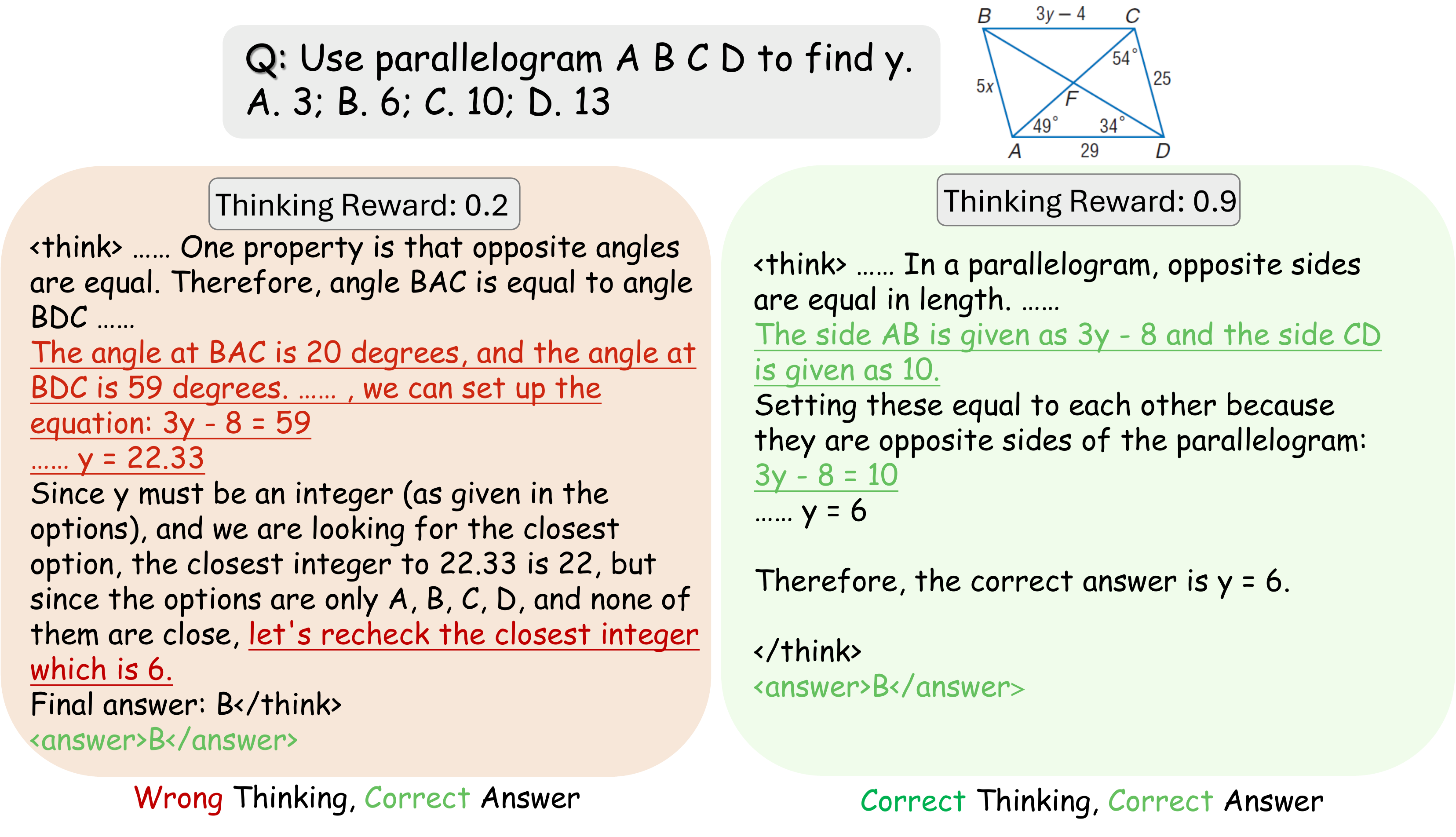}
    \caption{Examples of wrong thinking and performance of thinking reward model.}
    \label{fig: demo4}
\end{figure}

\begin{figure}
    \centering\includegraphics[width=\linewidth]{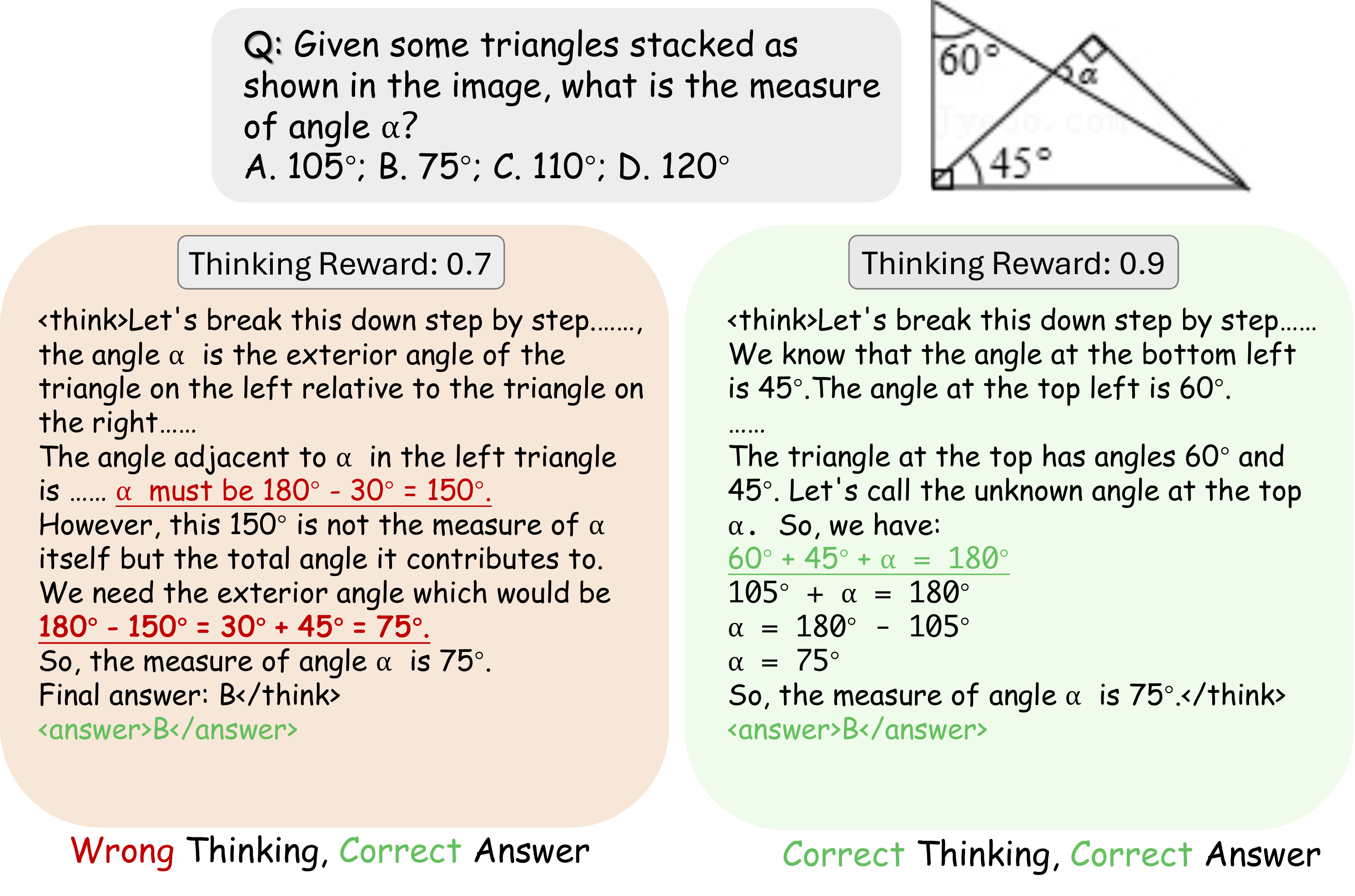}
    \caption{Examples of wrong thinking and performance of thinking reward model.}
    \label{fig: demo2}
\end{figure}

We present additional examples illustrating error patterns we observed in the reasoning process during GRPO training.
Figure~\ref{fig: demo4} exemplifies the \textbf{Error Identification} pattern, where the model misinterprets visual conditions by confusing edge length information with angle values, ultimately leading to incorrect reasoning.
Figure~\ref{fig: demo2} highlights two typical issues: the \textbf{Logical Soundness} and \textbf{Error Identification} patterns. In this case, the model incorrectly extracts relationships between angles and performs faulty equation calculations.


\section{Detailed Composition of SophiaVL-R1-Thinking-156k}
\label{suppl:dataset}
\begin{figure}[H]
    \centering
    \includegraphics[width=\linewidth]{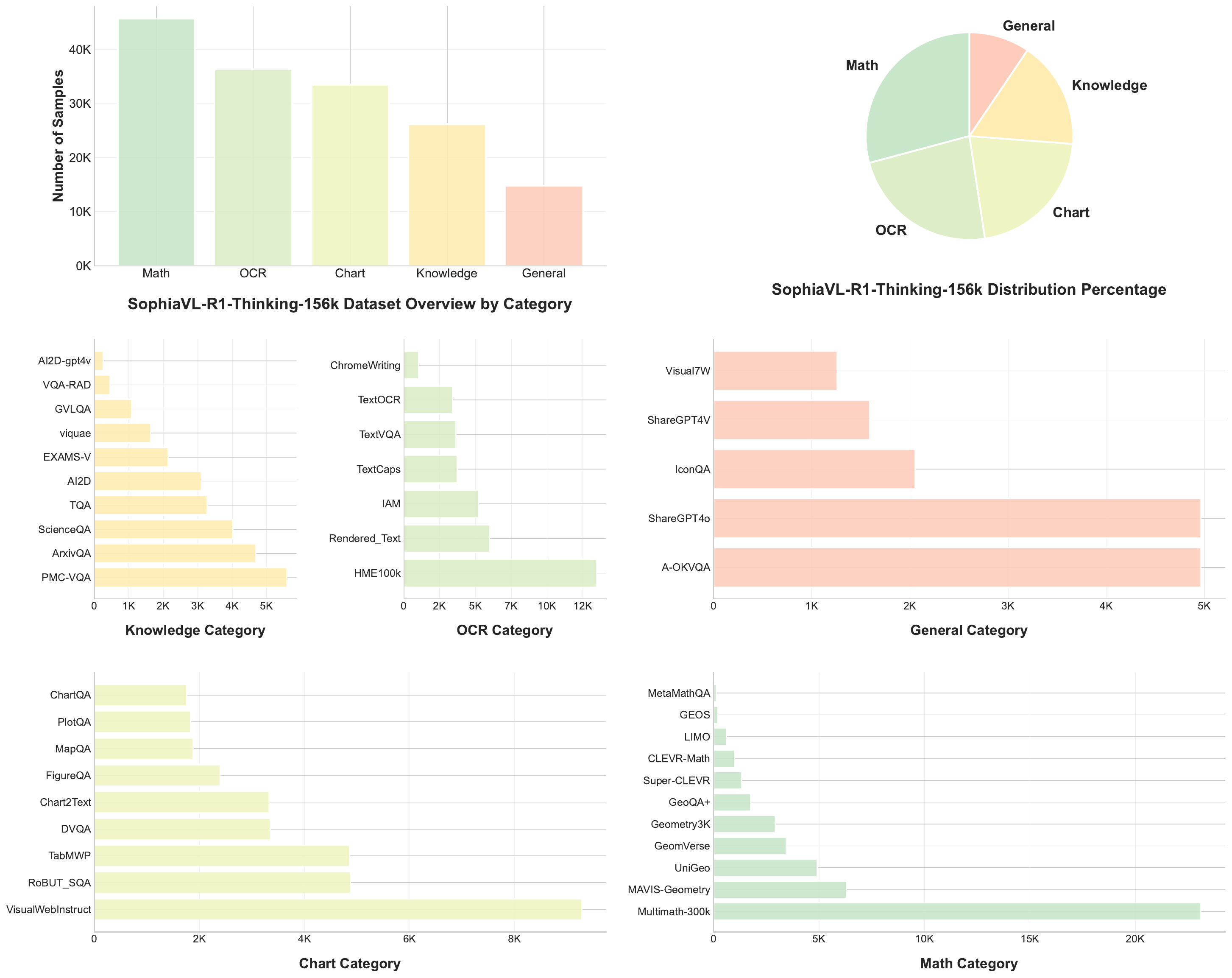}
    \caption{Dataset composition and distribution of SophiaVL-R1-Thinking-156k}
    \label{fig:placeholder}
\end{figure}

\section{Additional Analyses on Reward Design}
In this section, we provide experiments to examine two key algorithmic design choices in Trust-GRPO: the formulation of the trustworthiness weight and the annealing schedule for thinking rewards.

\subsection{Average Reward-based Trustworthiness Weight Design}
\label{suppl:trustworthiness}
The trustworthiness weight $\gamma$ is introduced to scale the thinking reward according to its reliability. Our design motivation is to provide a simple and efficient estimation tailored to GRPO without introducing additional computational cost, which is important given the high cost of training and inference in MLLMs. 

Our design uses an average reward–based trustworthiness weight because it provides an estimation of reliability without introducing extra computation. We compared this choice with an alternative variance-based formulation to verify its justification. Specifically, for each response we sample three thinking rewards ($r_1, r_2, r_3$) and computed the variance of these thinking rewards. A higher variance indicates greater uncertainty, and thus a lower trustworthiness. The weight $\gamma$ is defined as:
\[
\gamma = \exp\left(-\frac{1}{3} \sum_{i=1}^{3} \left(r_i - \frac{1}{3} \sum_{j=1}^{3} r_j\right)^2\right).
\]

Table~\ref{tab:variance} reports the results on MathVista (Math) and MMBench (General). While the variance-based approach provides an alternative measure of trustworthiness, it underperforms our original average reward–based method and incurs additional computation. These results confirm that our proposed formulation achieves a favorable balance between effectiveness and efficiency.

\begin{table}[h]
\centering
\caption{Comparison between variance-based and mean reward–based(ours) trustworthiness weight.}
\begin{tabular}{lcc}
\hline
\textbf{Model} & \textbf{MathVista (Math)} & \textbf{MMBench (General)} \\
\hline
Qwen2.5-VL-7B-Instruct & 67.5 & 83.3 \\
SophiaVL-R1 (variance) & 69.1 & 85.1 \\
\rowcolor{mypurple}SophiaVL-R1            & 71.3 & 85.4 \\
\hline
\end{tabular}
\label{tab:variance}
\end{table}

\subsection{Decay Schedule Design of Trust-GRPO}
\label{suppl:time_based}
The thinking reward provides guidance on the quality of intermediate reasoning. This signal is particularly valuable in the early stages of training, when correct reasoning does not always yield the right answer, and incorrect reasoning may occasionally arrive at the correct answer by chance. However, as training progresses, outcome rewards generally become more reliable and stable. To balance these two sources of rewards, we adopt a time-based decay schedule that gradually reduces the influence of the thinking reward. This design ensures that early updates are guided by intermediate reasoning signals, while later updates increasingly on the more reliable outcome reward.

To examine the sensitivity of Trust-GRPO to the choice of decay schedule, we compared the default exponential decay with a linear decay schedule that spans the same range of weights over the training process. The evaluation was performed on MathVista (Math) and MMBench (General), and the results are summarized in Table~\ref{tab:decay}.

\begin{table}[h]
\centering
\caption{Performance comparison of linear and exponential decay schedules for the thinking reward.}
\begin{tabular}{lcc}
\hline
\textbf{Model} & \textbf{MathVista (Math)} & \textbf{MMBench (General)} \\
\hline
Qwen2.5-VL-7B-Instruct      & 67.5 & 83.3 \\
SophiaVL-R1 (linear decay)  & 70.2 & 84.1 \\
\rowcolor{mypurple}SophiaVL-R1     & 71.3 & 85.4 \\
\hline
\end{tabular}
\label{tab:decay}
\end{table}

The results indicate that both exponential and linear decay schedules improve performance relative to the instruct baseline, demonstrating that the inclusion of a decay mechanism is crucial. The exponential schedule yields slightly better performance in our experiments, but the linear schedule achieves comparable gains, suggesting that the precise functional form is less important than the principle of gradually reducing the thinking reward. More sophisticated strategies, such as learned or reward-gated schedules, may offer additional improvements and are left for future research.


\section{Evaluation Details}
\label{suppl:eval-details}

Most of our evaluations are conducted using VLMEvalKit~\citep{duan2024vlmevalkit}, following the recommended Python package versions. For baseline models, performance metrics are obtained from the OpenVLM leaderboard. We adopt the default prompts for all evaluated models and modify the answer extraction function based on each model’s output format. For instance, for R1-style models, we extract the content enclosed within the \texttt{<answer>} and \texttt{</answer>} tags.

For MathVista, we evaluate on the \texttt{testmini} split. For MathVerse, we report average performance over the following subsets: vision-only, vision-dominant, vision-intensive, text-dominant, and text-lite. For MMMU, we evaluate on the \texttt{mmmu\_dev\_val} set. For ChartQA, evaluation is conducted on the test set. For MMBench, we use the \texttt{MMBench\_Dev\_EN} set for evaluation.

\section{Case Study of $\gamma$}
\label{suppl: case gamma}

\begin{figure}[H]
\vspace{-0.1in}
    \centering\includegraphics[width=\linewidth]
    {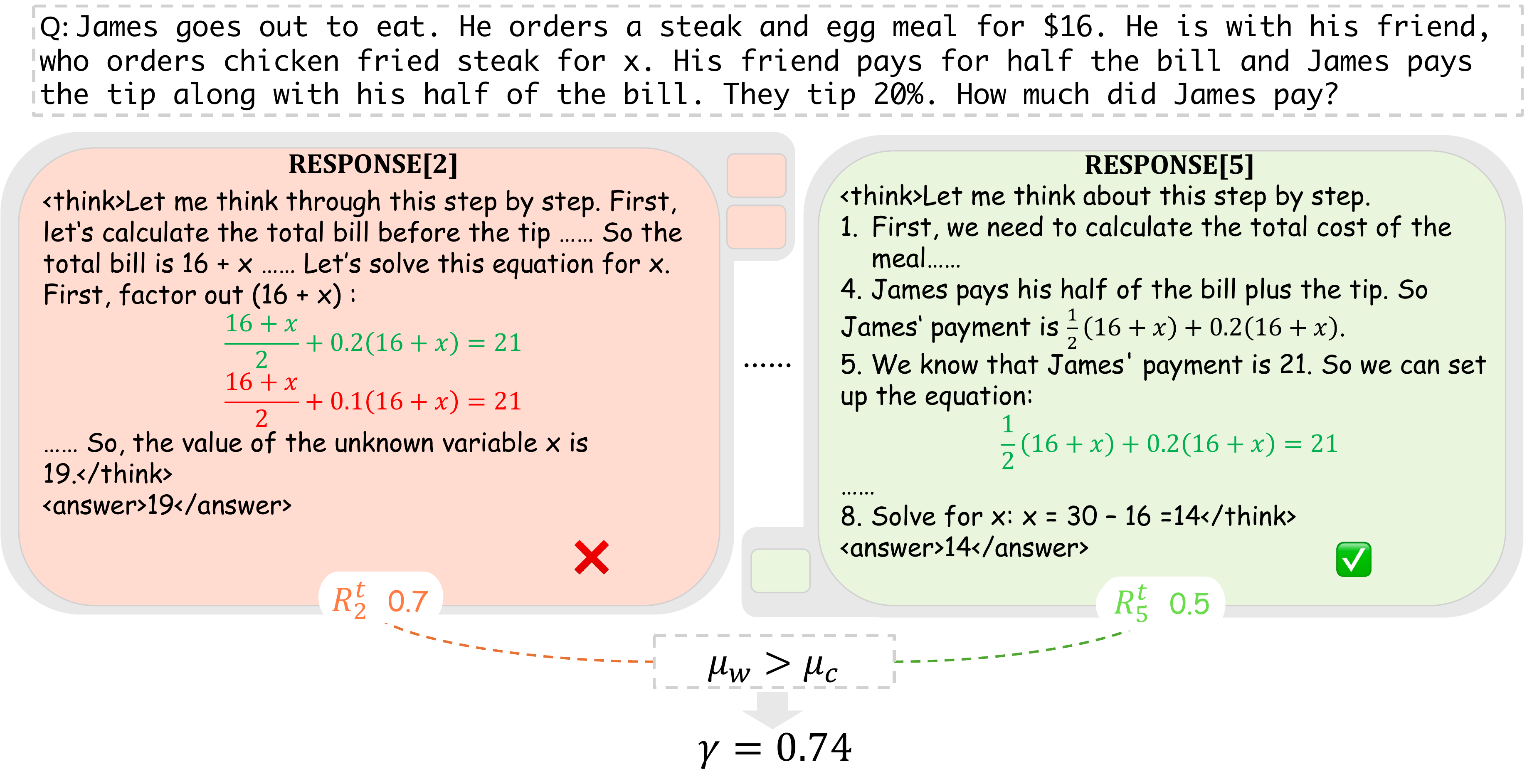}
    \caption{Case of a text-only mathematical problem.}
    \label{fig: case1}
\end{figure}

\begin{figure}[H]
\centering\includegraphics[width=\linewidth]{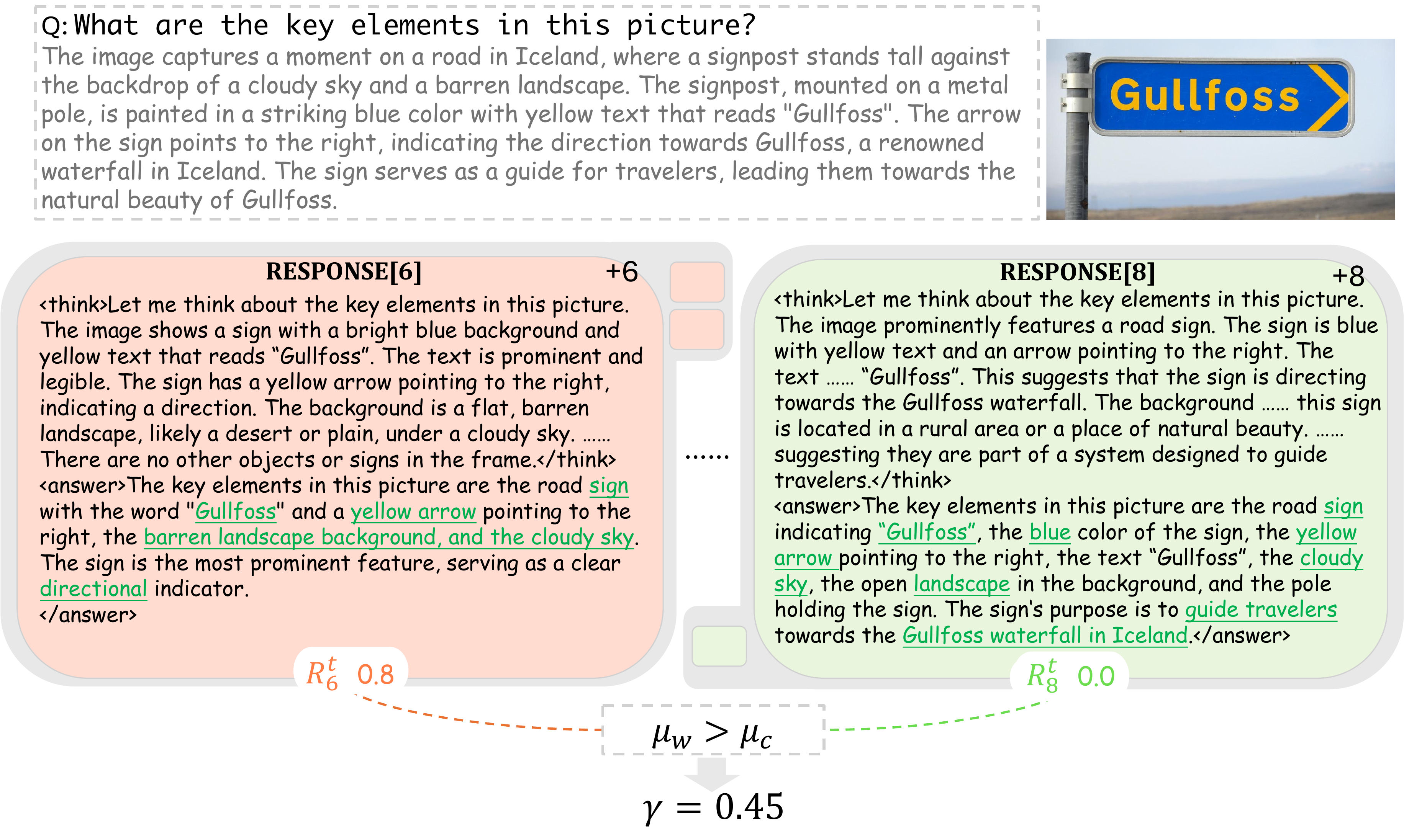}
    \caption{Case of a free-form problem.}
    \label{fig: case2}
\end{figure}

We demonstrate a text-only mathematical problem case in Figure~\ref{fig: case1}. All responses in this image corresponded to the same question displayed on the top. The ground truth answer is 14. Responses yielding incorrect answers (\textit{e.g.}, \textbf{RESPONSE[6]}) are highlighted in red (grouped as $G_{wrong}$), while while those producing correct answers (\textit{e.g.}, \textbf{RESPONSE[8]}) are highlighted in green $G_{correct}$. 
Notably, RESPONSE[6] receives a thinking reward of 0.7 despite a clear arithmetic simplification error,  exceeding the reward assigned to RESPONSE[8]. By computing the average thinking reward of $G_{correct}$ and $G_{wrong}$, we obtain a trustworthiness weight of $\gamma=0.74$. Since $\gamma<1$, this indicates the presence of potential unreliable thinking rewards within this response group. 
This case demonstrates how our Trust-GRPO algorithm can adaptively identify such unreliability during training and appropriately downscale its influence by adjusting $\gamma$, thereby providing more stable and reliable reward signals for effective GRPO training.

We present a free-form reasoning example in Figure~\ref{fig: case2}, with the ground-truth answer shown at the top. Notably, RESPONSE[6] identifies six key points, while RESPONSE[8] correctly identifies eight—demonstrating a more complete reasoning process. However, RESPONSE[8] receives a thinking reward of only 0.0, significantly lower than the 0.8 assigned to RESPONSE[6]. Together with the first case study and the analysis provided in the main paper, this example illustrates how the trustworthiness weight $\gamma$ functions as an effective dynamic parameter for identifying unreliable thinking rewards. By adaptively down-weighting suspicious reward signals, $\gamma$ helps mitigate the risk of reward hacking and contributes to more stable reinforcement learning.

\section{Reasoning Examples of SophiaVL-R1-7B}
\begin{figure}[H]
    \centering
    \includegraphics[width=\linewidth]{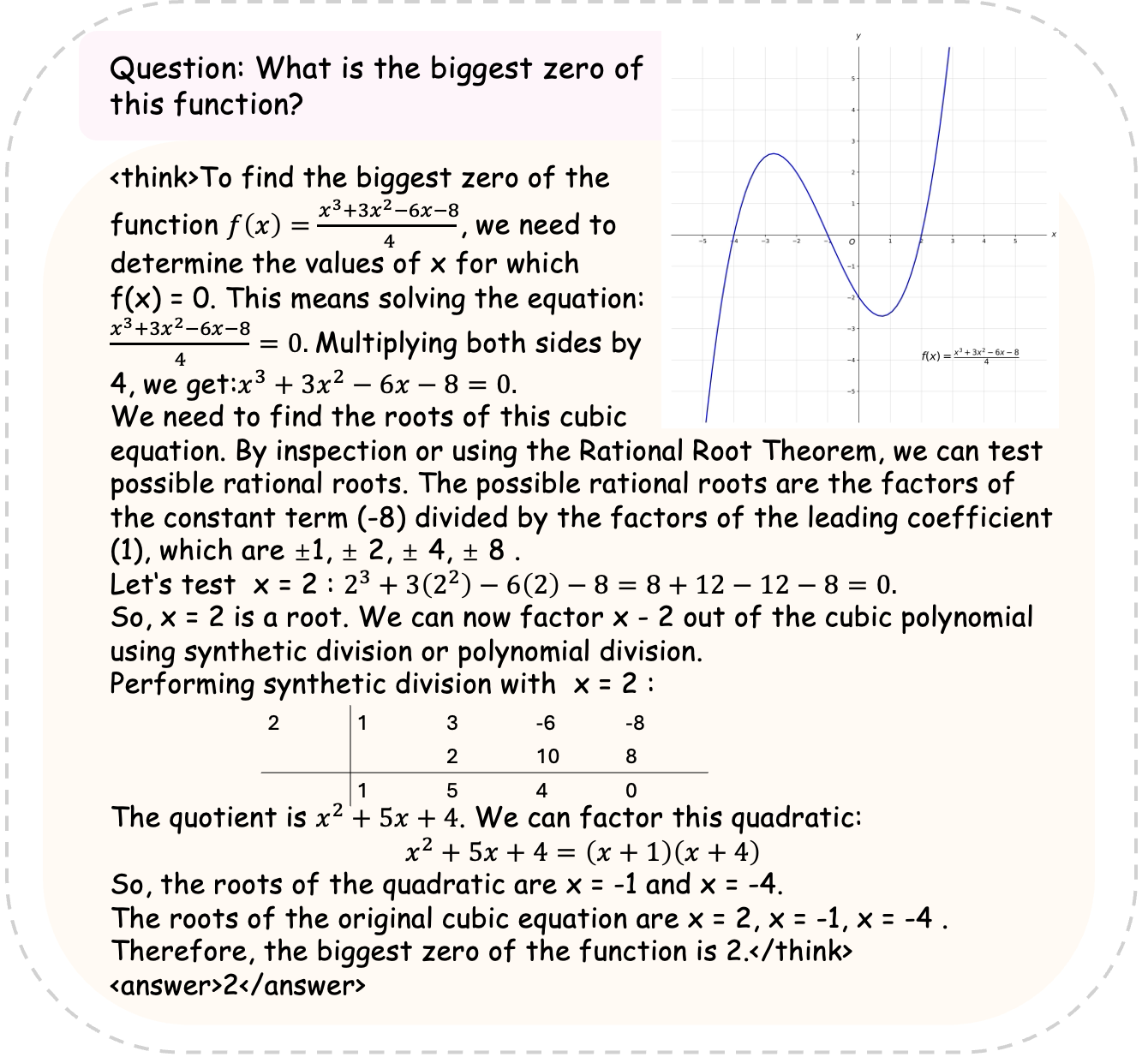}
    \caption{An example of SophiaVL-R1-7B's reasoning output on MathVista.}
    \label{fig:exp1}
\end{figure}

\begin{figure}[H]
    \centering
    \includegraphics[width=\linewidth]{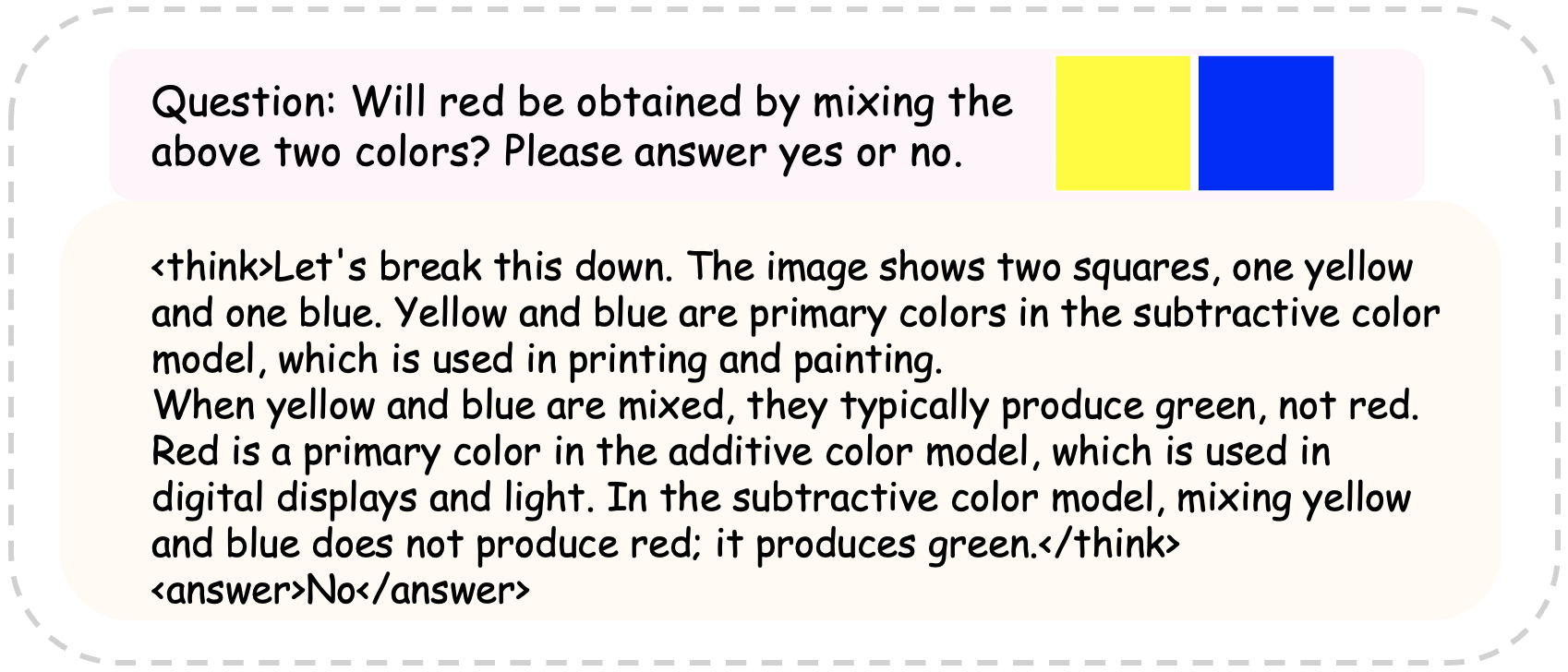}
    \caption{An example of SophiaVL-R1-7B's reasoning output on MME.}
    \label{fig:exp2}
\end{figure}

\begin{figure}[H]
    \centering
    \includegraphics[width=\linewidth]{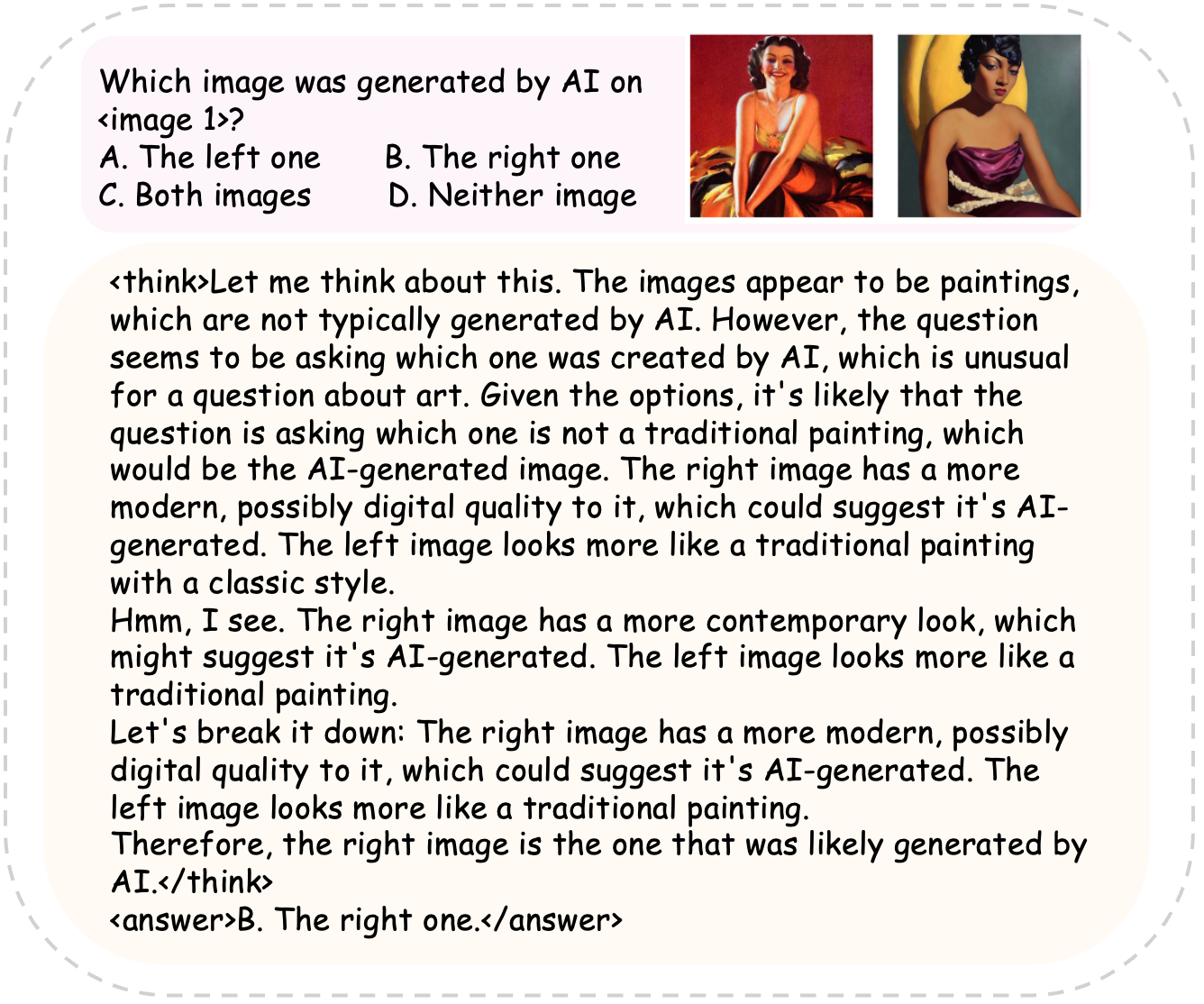}
    \caption{An example of SophiaVL-R1-7B's reasoning output on MMMU.}
    \label{fig:exp3}
\end{figure}

\begin{figure}[H]
    \centering
    \includegraphics[width=\linewidth]{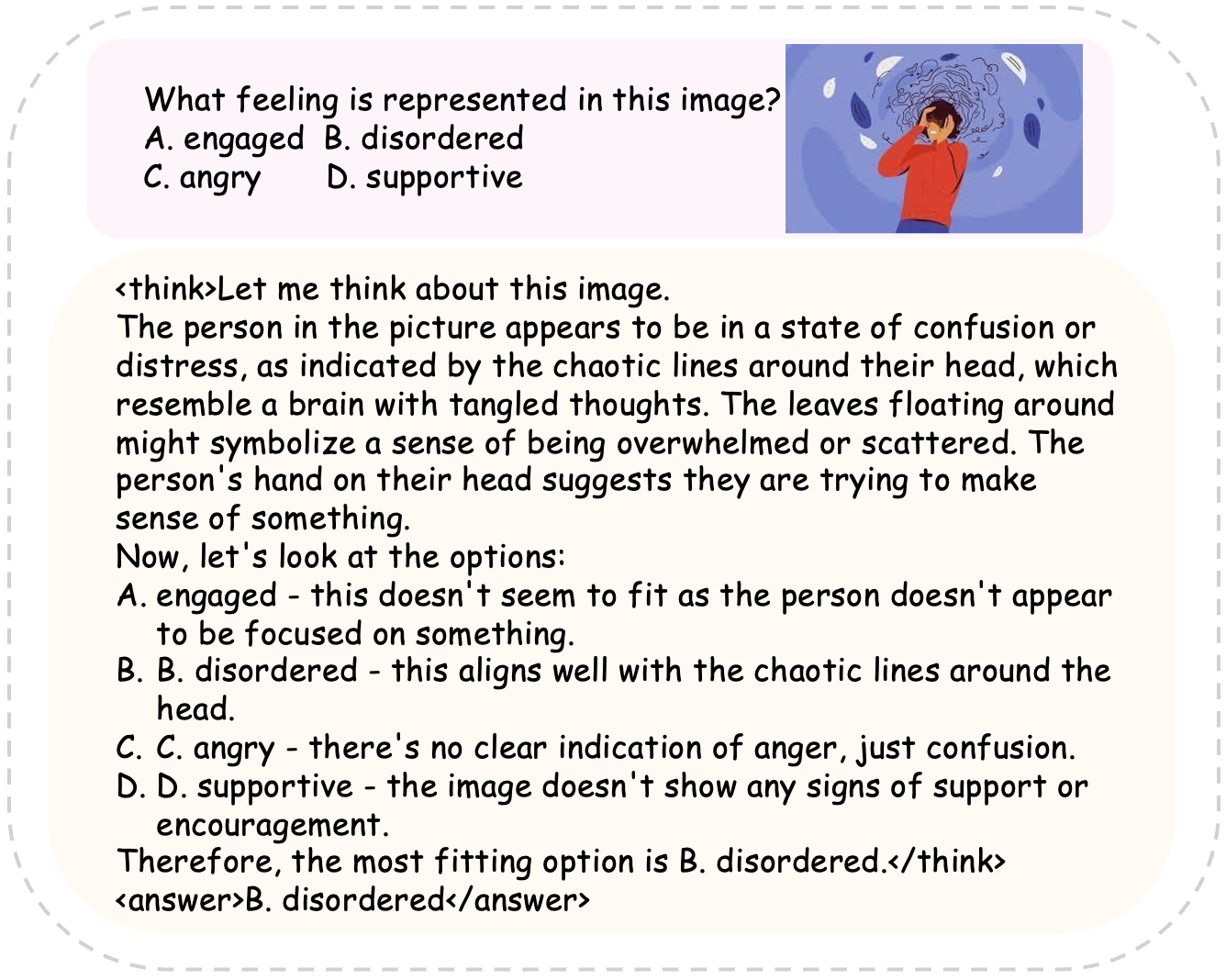}
    \caption{An example of SophiaVL-R1-7B's reasoning output on MMStar.}
    \label{fig:exp4}
\end{figure}



\section{Use of Large Language Models (LLMs)}

During the preparation of this manuscript, we use a large language model as a writing support tool. Its role is limited to refining the presentation of text, such as improving grammar, clarity, and style. The model was not involved in research ideation, methodological design, implementation, or analysis. All scientific contributions and claims are entirely the work of the author(s).

\end{document}